\documentclass[11pt,a4paper]{article}
\usepackage[hyperref]{acl2021}
\usepackage{times}
\usepackage{latexsym}

\usepackage[utf8]{inputenc} % allow utf-8 input
\usepackage[T1]{fontenc}    % use 8-bit T1 fonts
\usepackage{hyperref}       % hyperlinks
\usepackage{url}            % simple URL typesetting
\usepackage{booktabs}       % professional-quality tables
\usepackage{amsfonts}       % blackboard math symbols
\usepackage{nicefrac}       % compact symbols for 1/2, etc.
\usepackage{microtype}      % microtypography
\usepackage{tablefootnote}

\usepackage{microtype}

\aclfinalcopy % Uncomment this line for the final submission
\def\aclpaperid{1609} %  Enter the acl Paper ID here

\usepackage{multirow}
\usepackage{amsmath}
\usepackage{capt-of}
\usepackage{tabularx}
\usepackage[caption=false]{subfig}
\usepackage{epsfig}
\usepackage{amssymb}
\usepackage{scalerel}
\usepackage[inline]{enumitem}
\usepackage{listings}
\usepackage{varwidth}
\usepackage[export]{adjustbox}
\usepackage{tikz}
\usetikzlibrary{tikzmark}
\usepackage{todonotes}
\usepackage{cleveref}

\usepackage{stmaryrd}
\usepackage{bbm}

\usepackage{algorithm}
\usepackage[noend]{algpseudocode}

\definecolor{deepblue}{rgb}{0,0,0.5}
\definecolor{officeblue}{RGB}{0,102,204}
\definecolor{deepred}{rgb}{0.6,0,0}
\definecolor{deepgreen}{rgb}{0,0.5,0}
\definecolor{mybrickred}{RGB}{182,50,28}

\definecolor{fillcolor}{RGB}{216,217,252}

\algnewcommand\algorithmicrequireb{{\hspace{0.95cm}}}
\algnewcommand\INPTDESCB{\item[\algorithmicrequireb]}

\algnewcommand\algorithmicfuncdesc{\textbf{Function:}}
\algnewcommand\FUNCDESC{\item[\algorithmicfuncdesc]}
\algnewcommand\algorithmicfuncdescb{{\hspace{0.86cm}}}
\algnewcommand\FUNCDESCB{\item[\algorithmicfuncdescb]}
\algnewcommand{\algorithmicgoto}{\textbf{goto}}
\algnewcommand{\Goto}[1]{\algorithmicgoto~\ref{#1}}

\usepackage{amsmath,amsfonts,bm}

\def\eqref#1{equation~\ref{#1}}
\def\1{\bm{1}}

\DeclareMathAlphabet{\mathsfit}{\encodingdefault}{\sfdefault}{m}{sl}
\SetMathAlphabet{\mathsfit}{bold}{\encodingdefault}{\sfdefault}{bx}{n}

\newcommand{\softmax}{\mathrm{softmax}}

\newcommand\ours{Ours}
\newcommand\minilmvone{\textsc{MiniLM}}

\newcommand\bertsmall{BERT$_\text{SMALL}$}
\newcommand\bertbase{BERT$_\text{BASE}$}
\newcommand\bertlarge{BERT$_\text{LARGE}$}
\newcommand\bertlargewwm{BERT$_\text{LARGE-WWM}$}
\newcommand\xlmrbase{XLM-R$_\text{BASE}$}
\newcommand\xlmrlarge{XLM-R$_\text{LARGE}$}

\newcommand{\robertabase}{RoBERTa$_{\textsc{base}}$}
\newcommand{\robertalarge}{RoBERTa$_{\textsc{large}}$}

\title{\textsc{MiniLM}v2: Multi-Head Self-Attention Relation Distillation \\ for Compressing Pretrained Transformers}

\author{Wenhui Wang \quad Hangbo Bao \quad Shaohan Huang \quad Li Dong \quad Furu Wei\thanks{~~Contact person.} \\
Microsoft Research \\
\texttt{\{wenwan,t-habao,shaohanh,lidong1,fuwei\}@microsoft.com}}
\date{}

\begin{document}
\maketitle
\begin{abstract}
% In this work, we present a simple and effective dot-product self-attention distillation framework for task-agnostic compression of pretrained Transformers, especially for large-size models (24 layers, 1024 hidden size).
% We propose to transfer self-attention relations, which are computed by multi-head dot-product of different combinations of queries, keys and values in self-attention module (e.g., query-query and query-value relation). 
% The number of relation heads can be different from the attention head number of teacher and student models, which avoids the limitation on the attention head number of student models introduced by transferring teacher's self-attention distributions (attention maps/weights).
% Inspired by \minilmvone{}~\cite{minilmv1}, we use knowledge of one layer of the teacher to train the student.
% For the large-size model, we show that transferring self-attention relations of an upper middle layer of the teacher model achieves better performance than other layers. 
% Experimental results demonstrate that our models\footnote{Models and code will be publicly available at \url{https://aka.ms/minilm}.} distilled from different base-size and large-size teachers (BERT and RoBERTa) outperform state-of-the-art baselines. 
% The 6-layer 768-dimensional model distilled from \bertlarge{} is 2.0$\times$ faster than \bertbase{}, while archiving better performance on SQuAD 2.0 and GLUE benchmark. 
% The base-size model distilled from \robertalarge{} also achieves competitive performance.

We generalize deep self-attention distillation in \minilmvone{}~\cite{minilmv1} by only using self-attention relation distillation for task-agnostic compression of pretrained Transformers.
In particular, we define multi-head self-attention relations as scaled dot-product between the pairs of \textit{query}, \textit{key}, and \textit{value} vectors within each self-attention module.
Then we employ the above relational knowledge to train the student model.
Besides its simplicity and unified principle, more favorably, there is no restriction in terms of the number of student's attention heads, while most previous work has to guarantee the same head number between teacher and student.
% which is required to be the same as its teacher for  transferring self-attention distributions (attention maps).
Moreover, the fine-grained self-attention relations tend to fully exploit the interaction knowledge learned by Transformer.
% using more relation heads in assessing the  brings more  self-attention knowledge and improves the performance of the student model.
%In addition to base-size (12 layers, 768 hidden size) model, we conduct extensive distillation experiments on large-size teacher model (24 layers, 1024 hidden size) and find that
In addition, we thoroughly examine the layer selection strategy for teacher models, rather than just relying on the last layer as in \minilmvone{}.
% find that transferring self-attention relations of an upper middle layer of the large-size teacher model achieves better performance than using the last layer as in \minilmvone{}. 
We conduct extensive experiments on compressing both monolingual and multilingual pretrained models.
Experimental results demonstrate that our models\footnote{Distilled models and code will be publicly available at \url{https://aka.ms/minilm}.} distilled from base-size and large-size teachers (BERT, RoBERTa and XLM-R) outperform the state-of-the-art.
%The 6-layer 768-dimensional model distilled from \bertlarge{} is 2.0$\times$ faster than \bertbase{}, while archiving better performance on SQuAD 2.0 and GLUE benchmark. 
%The base-size model distilled from \robertalarge{} also achieves competitive performance.

% we present a simple and effective dot-product self-attention distillation framework for task-agnostic compression of pretrained Transformers, especially for large-size models (24 layers, 1024 hidden size).

% Dot-product of queries-queries, keys-keys and values-values of teacher's self-attention module are utilized to train student models. Using these dot-product knowledge instead of the attention distributions (dot-produce of queries and keys) avoids the limit on the number of attention head of student models. 
% More importantly, for the large-size model, we show that transferring the self-attention knowledge of a upper middle layer of the teacher model achieves better performance than other layers. 
% Experimental results demonstrate that our models distilled from different base-size and large-size teachers (BERT and RoBERTa) outperform state-of-the-art baselines in different parameter size. 
% Our base-size model distilled from \robertalarge{} outperforms \robertabase{} while having a lower computational cost.

\end{abstract}

\section{Introduction}

Pretrained Transformers~\cite{gpt,bert,unilm,xlnet,spanbert,roberta,unilmv2,gpt2,t5,bart} have been highly successful for a wide range of natural language processing tasks. 
However, these models usually consist of hundreds of millions of parameters and are getting bigger. 
It brings challenges for fine-tuning and online serving in real-life applications due to the restrictions of computation resources and latency.

Knowledge distillation (KD;~\citealt{softlabeldistill},~\citealt{intermediatedistill}) has been widely employed to compress pretrained Transformers, which transfers knowledge of the large model (teacher) to the small model (student) by minimizing the differences between teacher and student features. 
Soft target probabilities (soft labels) and intermediate representations are usually utilized to perform KD training.
In this work, we focus on task-agnostic compression of pretrained Transformers~\cite{distillbert,mminibert,tinybert,mobilebert,minilmv1}. The student models are distilled from large pretrained Transformers using large-scale text corpora.
The distilled task-agnostic model can be directly fine-tuned on downstream tasks, and can be utilized to initialize task-specific distillation.
% Task-specific distillation firstly fine-tune the large model on specific task and then transfer its knowledge to the student model using task data. 
% Compared with task-specific distillation, task-agnostic distillation is more deniable. Moreover, the distilled task-agnostic model can be utilized to initialize task-specific distillation to further improve its performance. 
% Intermediate representations (such as self-attention distributions and hidden states) and soft target probabilities (soft label) for masked language modeling predictions are usually used to guide the training of the student model. 

DistilBERT~\cite{distillbert} uses soft target probabilities for masked language modeling predictions and embedding outputs to train the student. The student model is initialized from the teacher by taking one layer out of two.
TinyBERT~\cite{tinybert} utilizes hidden states and self-attention distributions (i.e., attention maps and weights), and adopts a uniform function to map student and teacher layers for layer-wise distillation. 
MobileBERT~\cite{mobilebert} introduces specially designed teacher and student models using inverted-bottleneck and bottleneck structures to keep their layer number and hidden size the same, layer-wisely transferring hidden states and self-attention distributions. 
\minilmvone{}~\cite{minilmv1} proposes deep self-attention distillation, which uses self-attention distributions and value relations to help the student to deeply mimic teacher's self-attention modules.
% by deeply mimicking teacher's self-attention module.
% to deeply mimic teacher's self-attention module.
% also uses self-attention distributions and further introduces value relation to train the student by deeply mimicking teacher's self-attention module. 
% Different from previous work,
\minilmvone{} shows that transferring knowledge of teacher's last layer achieves better performance than layer-wise distillation.
In summary, most previous work relies on self-attention distributions to perform KD training, which leads to a restriction that the number of attention heads of student model has to be the same as its teacher.

% The student model is required to have the same number of attention heads as its teacher. 
% Besides, in addition to MobileBERT which uses a specially designed large-size teacher model, most of previous work conduct experiments using a base-size teacher model. 

In this work, we generalize and simplify deep self-attention distillation of \minilmvone{}~\cite{minilmv1} by using self-attention relation distillation.
We introduce multi-head self-attention relations computed by scaled dot-product of pairs of queries, keys and values, which guides the student training.
Taking query vectors as an example, in order to obtain queries of multiple relation heads, we first concatenate query vectors of different attention heads, and then split the concatenated vector according to the desired number of relation heads.
% The number of relation heads can be different from the number of attention heads of teacher and student models, which allows more flexibility for the number of student's attention heads.
Afterwards, for teacher and student models with different attention head numbers, we can align their queries with the same number of relation heads for distillation.
% attention head number of teacher and student models, and therefore the student can use arbitrary number of attention heads.
Moreover, using a larger number of relation heads brings more fine-grained self-attention knowledge, which helps the student to achieves a deeper mimicry of teacher's self-attention module.
In addition, for large-size ($24$ layers, $1024$ hidden size) teachers, extensive experiments indicate that transferring an upper middle layer tends to perform better than using the last layer as in \minilmvone{}.

Experimental results show that our monolingual models distilled from BERT and RoBERTa, and multilingual models distilled from XLM-R outperform state-of-the-art models in different parameter sizes.
% Our method is effective for different base-size and large-size teachers.
The $6$$\times$$768$ ($6$ layers, $768$ hidden size) model distilled from \bertlarge{} is $2.0\times$ faster, meanwhile, performing better than \bertbase{}. The base-size model distilled from \robertalarge{} outperforms \robertabase{} using much fewer training examples.
% Experimental results demonstrate that our method is effective for different base-size and large-size teachers.
% Moreover, our multilingual models distilled from \xlmrbase{} and \xlmrlarge{} also obtain competitive results.

To summarize, our contributions include:
\begin{itemize}
\item We generalize and simplify deep self-attention distillation in \minilmvone{} by introducing multi-head self-attention relation distillation, which brings more fine-grained self-attention knowledge and allows more flexibility for the number of student's attention heads.
\item We conduct extensive distillation experiments on different large-size teachers and find that using knowledge of a teacher's upper middle layer achieves better performance.
\item Experimental results demonstrate the effectiveness of our method for different monolingual and multilingual teachers in base-size and large-size.
\end{itemize}

\begin{figure*}[t]
\centering
\includegraphics[width=0.98\textwidth]{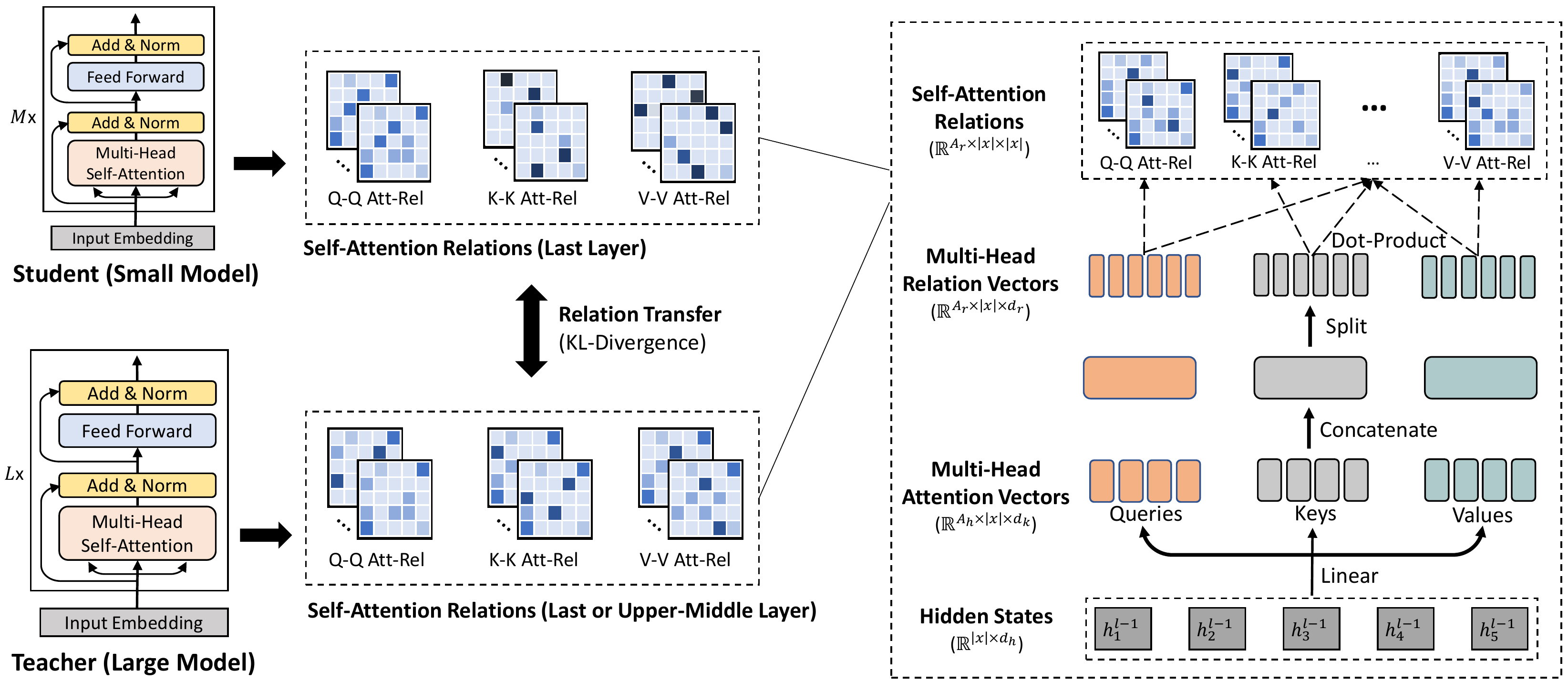}
\caption{Overview of multi-head self-attention relation distillation. 
We introduce multi-head self-attention relations computed by scaled dot-product of pairs of queries, keys and values to guide the training of students.
In order to obtain vectors (queries, keys and values) of multiple relation heads, we first concatenate self-attention vectors of different attention heads and then split them according to the desired number of relation heads.
We choose to transfer Q-Q, K-K and V-V self-attention relations to achieve a balance between performance and training speed.
For large-size teacher, we transfer the self-attention knowledge of an upper middle layer of the teacher. 
For base-size teacher, using the last layer achieves better performance. 
% Our student models are named as \ours{}.
}
\label{fig:method}
\end{figure*}

\begin{table*}[t]
% \vskip 0.1in
\begin{center}
\scalebox{0.67}{
\begin{tabular}{l|c|cc|c|ccccccc|c}
\toprule
\bf Model & \bf Teacher & \bf \#Param & \bf Speedup & \bf SQuAD2 & \bf MNLI-m & \bf QNLI & \bf QQP & \bf RTE & \bf SST & \bf MRPC & \bf CoLA & \bf Avg \\ 
\midrule
\bertbase{} & - & 109M & $\times$1.0 & 76.8 & 84.5 & 91.7 & 91.3 & 68.6 & 93.2 & 87.3 & 58.9 & 81.5 \\
\robertabase{} & - & 125M & $\times$1.0 & 83.7 & 87.6 & 92.8 & 91.9 & 78.7 & 94.8 & 90.2 & 63.6 & 85.4 \\
% \bertlarge{}~\cite{bert} & 109M & 76.8 & 84.5 & 93.2 & 91.7 & 58.9 & 68.6 & 87.3 & 91.3 & 81.5 \\
\midrule
\bertsmall{} & - & 66M & $\times$2.0 & 73.2 & 81.8 & 89.8 & 90.6 & 67.9 & 91.2 & 84.9 & 53.5 & 79.1 \\
Truncated \bertbase{} & - & 66M & $\times$2.0 & 69.9 & 81.2 & 87.9 & 90.4 & 65.5 & 90.8 & 82.7 & 41.4 & 76.2 \\
Truncated \robertabase{} & - & 81M & $\times$2.0 & 77.9 & 84.9 & 91.1 & 91.3 & 67.9 & 92.9 & 87.5 & 55.2 & 81.1 \\
DistilBERT & \bertbase{} & 66M & $\times$2.0 & 70.7 & 82.2 & 89.2 & 88.5 & 59.9 & 91.3 & 87.5 & 51.3 & 77.6 \\
TinyBERT & \bertbase{} & 66M & $\times$2.0 & 73.1 & 83.5 & 90.5 & 90.6 & 72.2 & 91.6 & 88.4 & 42.8 & 79.1 \\
\textsc{MiniLM} & \bertbase{} & 66M & $\times$2.0 & 76.4  & 84.0 & 91.0 & 91.0 & 71.5 & 92.0 & 88.4 & 49.2 & 80.4 \\
\midrule
$6$$\times$$384$ Ours & \bertbase{} & 22M & $\times$5.3 & 72.9 & 82.8 & 90.3 & 90.6 & 68.9 & 91.3 & 86.6 & 41.8 & 78.2 \\
% 6$\times$384 Ours$\dag$ & \bertbase{} & 22M & $\times$5.3 & 73.3 & 82.7 & 90.3 & 90.4 & 70.9 & 91.6 & 86.7 & 42.8 & 78.6 \\
$6$$\times$$384$ Ours & \bertlarge{} & 22M & $\times$5.3 & 74.3  & 83.0 & 90.4 & 90.7 & 68.5 & 91.1 & 87.8 & 41.6 & 78.4 \\
% 6$\times$384 Ours$\dag$ & \bertlarge{} & 22M & $\times$5.3 & 74.7  & 83.2 & 90.4 & 90.6 & 69.3 & 92.4 & 87.8 & 41.9 & 78.8 \\
$6$$\times$$384$ Ours & \robertalarge{} & 30M & $\times$5.3 & \textbf{76.4}  & \textbf{84.4} & \textbf{90.9} & \textbf{90.8} & \textbf{69.9} & \textbf{92.0} & \textbf{88.7} & \textbf{42.6} & \textbf{79.5} \\
\midrule
$6$$\times$$768$ Ours & \bertbase{} & 66M & $\times$2.0 & 76.3  & 84.2 & 90.8 & 91.1 & 72.1 & 92.4 & 88.9 & 52.5 & 81.0 \\
% 6$\times$768 \ours{}$\dag$ & \bertbase{} & 66M & $\times$2.0 & 76.8 & 84.4 & 91.1 & 91.1 & 73.7 & 92.3 & 88.6 & 54.7 & 81.6 \\
$6$$\times$$768$ Ours & \bertlarge{} & 66M & $\times$2.0 & 77.7  & 85.0 & 91.4 & 91.1 & 73.0 & 92.5 & 88.9 & 53.9 & 81.7 \\
% 6$\times$768 Ours$\dag$ & \bertlarge{} & 66M & $\times$2.0 & 78.1  & 85.2 & 91.3 & 91.1 & 73.3 & 92.5 & 88.5 & 54.4 & 81.8 \\
$6$$\times$$768$ Ours & \robertalarge{} & 81M & $\times$2.0 & \textbf{81.6}  & \textbf{87.0} & \textbf{92.7} & \textbf{91.4} & \textbf{78.7} & \textbf{94.5} & \textbf{90.4} & \textbf{54.0} & \textbf{83.8} \\
% \midrule
% 12$\times$384 Ours & \robertalarge{} & 33M & $\times$2.7 & \textbf{82.3}  & \textbf{86.9} & \textbf{92.7} & \textbf{91.3} & \textbf{77.6} & \textbf{93.5} & \textbf{90.0} & \textbf{53.8} & \textbf{83.5} \\
\bottomrule
\end{tabular}
}
\end{center}
\caption{
Results of our students distilled from base-size and large-size teachers on the development sets of GLUE and SQuAD 2.0. 
% Comparison between the publicly released $6$-layer models with 768 hidden size distilled from \bertbase{}.
% We compare task-agnostic distilled models without task-specific distillation and data augmentation.
We report F1 for SQuAD 2.0, Matthews correlation coefficient for CoLA, and accuracy for other datasets. 
The GLUE results of DistilBERT are taken from~\citet{distillbert}. 
The rest results of DistilBERT, TinyBERT\footnotemark, \bertsmall{}, Truncated \bertbase{} and \minilmvone{} are taken from~\citet{minilmv1}.
\bertsmall{}~\cite{studentInit} is trained using the MLM objective, without using KD.
We also report the results of truncated \bertbase{} and truncated \robertabase{}, which drops the top $6$ layers of the base model.
Top-layer dropping has been proven to be a strong baseline~\cite{poormanbert}. 
The fine-tuning results are an average of $4$ runs.
% We report the SQuAD 2.0 result by fine-tuning their released model$^{5}$.
% For TinyBERT, we fine-tune the latest version of their public model$^{6}$ for a fair comparison. We also report the fine-tuning results of Truncated \bertbase{} and the $6$x$768$ BERT model (\bertsmall{})~\cite{studentInit} trained using the MLM objective as baselines. \citet{poormanbert} show that top-layer dropping consistently outperforms other strategies when
% dropping 6 layers, so we drop top 6 layers from \bertbase{} for truncated \bertbase{}. 
}
% \vskip -0.1in
\label{tbl:main_exps}
\end{table*}

\section{Related Work}

\subsection{Backbone Network: Transformer}

Multi-layer Transformer~\cite{transformer} has been widely adopted in pretrained models. 
Each Transformer layer consists of a self-attention sub-layer and a position-wise fully connected feed-forward sub-layer.

\paragraph{Self-Attention}

Transformer relies on multi-head self-attention to capture dependencies between words. Given previous Transformer layer's output $\mathbf{H}^{l-1} \in \mathbb{R}^{|x| \times d_h}$, the output of a self-attention head $\mathbf{O}_{l,a},~a \in [1, A_h]$ is computed via:
\begin{gather}
\mathbf{Q}_{l,a} = \mathbf{H}^{l-1} \mathbf{W}_{l,a}^Q \\
\mathbf{K}_{l,a} = \mathbf{H}^{l-1} \mathbf{W}_{l,a}^K \\ 
\mathbf{V}_{l,a} = \mathbf{H}^{l-1} \mathbf{W}_{l,a}^V \\
% \mathbf{A}_{l,a} = \softmax(\frac{\mathbf{Q}_{l,a} \mathbf{K}_{l,a}^{\intercal}}{ \sqrt{d_k}})  \\
\mathbf{O}_{l,a} = \softmax(\frac{\mathbf{Q}_{l,a} \mathbf{K}_{l,a}^{\intercal}}{ \sqrt{d_k}})\mathbf{V}_{l,a}
\end{gather}
Previous layer's output is linearly projected to queries, keys and values using parameter matrices $\mathbf{W}_{l,a}^Q , \mathbf{W}_{l,a}^K , \mathbf{W}_{l,a}^V \in \mathbb{R}^{d_h \times d_k}$, respectively. The self-attention distributions are computed via scaled dot-product of queries and keys. These weights are assigned to the corresponding value vectors to obtain the attention output.
$|x|$ represents the length of input sequence. $A_h$ and $d_h$ indicate the number of attention heads and hidden size. $d_k$ is the attention head size. 
$d_k$$\times$$A_h$ is usually equal to $d_h$.

\subsection{Pretrained Language Models}

Pre-training has led to strong improvements across a variety of natural language processing tasks. 
Pretrained language models are learned on large amounts of text data, and then fine-tuned to adapt to specific tasks. 
BERT~\cite{bert} proposes to pretrain a deep bidirectional Transformer using masked language modeling (MLM) objective. \textsc{UniLM}~\cite{unilm} is jointly pretrained on three types language modeling objectives to adapt to both understanding and generation tasks.
% XLNet~\cite{xlnet} introduces permutation language modeling objective to predict masked tokens auto-regressively. 
% SpanBERT~\cite{spanbert} improves BERT by incorporating span information. 
RoBERTa~\cite{roberta} achieves strong performance by training longer steps using large batch size and more text data.  
MASS~\cite{mass}, T5~\cite{t5} and BART~\cite{bart} employ a standard encoder-decoder structure and pretrain the decoder auto-regressively. 
% \citet{unilmv2} propose a pseud-masked language model by jointly pretrained on MLM and partially auto-regressive MLM objectives.
Besides monolingual pretrained models, multilingual pretrained models~\cite{bert,xlm,xnlg,xlmr,infoxlm} also advance the state-of-the-art on cross-lingual understanding and generation.

\subsection{Knowledge Distillation}

Knowledge distillation has been proven to be a promising way to compress large models while maintaining accuracy. Knowledge of a single or an ensemble of large models is used to guide the training of small models.
\citet{softlabeldistill} propose to use soft target probabilities to train student models.
% \citet{intermediatedistill} introduce intermediate representations from hidden layers to train the student.   
More fine-grained knowledge such as hidden states~\cite{intermediatedistill} and attention distributions~\cite{att_dst,att_dst_mrc} are introduced to improve the student model. 

In this work, we focus on task-agnostic knowledge distillation of pretrained Transformers. 
The distilled task-agnostic model can be fine-tuned to adapt to downstream tasks. 
It can also be utilized to initialize task-specific distillation~\cite{patientdistill,studentInit,bertintermediate,xtremedistil,bertoftheseus,dynabert,bertemd}, which uses a fine-tuned teacher model to guide the training of the student on specific tasks.
Knowledge used for distillation and layer mapping function are two key points for task-agnostic distillation of pretrained Transformers. Most previous work uses soft target probabilities, hidden states, self-attention distributions and value-relation to train the student model. 
For the layer mapping function, TinyBERT~\cite{tinybert} uses a uniform strategy to map teacher and student layers. MobileBERT~\cite{mobilebert} assumes the student has the same number of layers as its teacher to perform layer-wise distillation. \minilmvone{}~\cite{minilmv1} transfers self-attention knowledge of teacher's last layer to the student last Transformer layer. Different from previous work, our method uses multi-head self-attention relations to eliminate the restriction on the number of student's attention heads. Moreover, we show that transferring the self-attention knowledge of an upper middle layer of the large-size teacher model is more effective. 

% first propose transferring the knowledge of the teacher to the student by using its soft target distributions to train the distilled model. \citet{intermediatedistill} introduce intermediate representations from hidden layers of the teacher to guide the training of the student. Knowledge of the attention maps~\cite{att_dst,att_dst_mrc} is also introduced to help the training.

\section{Multi-Head Self-Attention Relation Distillation}

Following \minilmvone{}~\cite{minilmv1}, the key idea of our approach is to deeply mimic teacher's self-attention module, which draws dependencies between words and is the vital component of Transformer.
\minilmvone{} uses teacher's self-attention distributions to train the student model. 
It brings the restriction on the number of attention heads of students, which is required to be the same as its teacher. 
To introduce more fine-grained self-attention knowledge and avoid using teacher's self-attention distributions, we generalize deep self-attention distillation in \minilmvone{} and introduce multi-head self-attention relations of pairs of queries, keys and values to train the student.
Besides, we conduct extensive experiments and find that layer selection of the teacher model is critical for distilling large-size models.
Figure~\ref{fig:method} gives an overview of our method.

\footnotetext{In addition to task-agnostic distillation, TinyBERT uses task-specific distillation and data augmentation to further improve the model. We report the fine-tuning results of their public task-agnostic model.}

% The relation is also computed via the multi-head scaled dot-product, but the head number used in relation can be more flexible. It can be different from the attention head number of teacher and student models. 
% Moreover, we find that using a larger number of relation heads can achieve better performance.

% The key idea of method is to deeply mimic teacher's self-attention module, which is the vital component in Transformer and draws dependencies between words. 
% In self-attention module, queries, key and values are the most basic and important vectors. To achieve a deeper mimicry and avoid using teacher's attention distributions with a fix attention head number, we introduce relations between queries, keys and values to perform knowledge distillation. The relation is also computed via the multi-head scaled dot-product, but we show that the head number used in relation can be different from the attention head number. Beside, using a larger number of relation heads can achieve better performance.

\begin{table*}[t]
% \vskip 0.1in
\begin{center}
\scalebox{0.65}{
\begin{tabular}{l|c|cc|c|cccccccc|c}
\toprule
\bf Model & \bf  Teacher & \bf \#Param & \bf Speedup & \bf SQuAD2 & \bf MNLI-m/mm & \bf QNLI & \bf QQP & \bf RTE & \bf SST & \bf MRPC & \bf CoLA & \bf STS & \bf Avg \\ 
\midrule
\bertbase{} & - & 109M & 1.0$\times$ & 76.8 & 84.6/83.4 & 90.5 & 71.2 & 66.4 & 93.5 & 88.9 & 52.1 & 85.8 & 79.3 \\
\bertlarge{} & - & 340M & 0.3$\times$ & 81.9 & 86.7/85.9 & 92.7 & 72.1 & 70.1 & 94.9 & 89.3 & 60.5 & 86.5 & 82.1 \\
\midrule
% MobileBERT & IB-\bertlarge{} & 25M & 1.8$\times$ & 80.2 & 84.3/83.4 & 91.6 & 70.5 & 70.4 & 92.6 & 88.8 & \textbf{51.1} & 84.8 & 79.8 \\
$6$$\times$$768$ \ours{} & \bertbase{} & 66M & 2.0$\times$ & 76.3 & 83.8/83.3 & 90.2 & 70.9 & \textbf{69.2} & 92.9 & \textbf{89.1} & 46.6 & 84.3 & 78.7 \\
$6$$\times$$768$ \ours{} & \bertlarge{} & 66M & 2.0$\times$ & \textbf{77.7} & \textbf{84.5}/\textbf{84.0} & \textbf{91.5} & \textbf{71.3} & \textbf{69.2} & \textbf{93.0} & \textbf{89.1} & \textbf{48.6} & \textbf{85.1} & \textbf{79.4} \\
\bottomrule
\end{tabular}
}
\end{center}
\caption{
Results of our $6$$\times$$768$ students distilled form BERT on GLUE test sets and SQuAD 2.0 dev set. The reported results are directly fine-tuned on downstream tasks. We report F1 for SQuAD 2.0, QQP and MRPC, Spearman correlation for STS-B, Matthews correlation coefficient for CoLA and accuracy for the rest.
}
% \vskip -0.1in
\label{tbl:6l_768_test_exps}
\end{table*}

\subsection{Multi-Head Self-Attention Relations}

Multi-head self-attention relations are obtained by scaled dot-product of pairs\footnote{There are nine types of self-attention relations, such as query-query, key-key, key-value and query-value relations.} of queries, keys and values of multiple relation heads.
Taking query vectors as an example, in order to obtain queries of multiple relation heads,
we first concatenate queries of different attention heads and then split the concatenated vector based on the desired number of relation heads. 
The same operation is also performed on keys and values.
For teacher and student models which uses different number of attention heads, we convert their queries, keys and values of different number of attention heads into vectors of the same number of relation heads to perform KD training.
Our method eliminates the restriction on the number of attention heads of student models.
Moreover, using more relation heads in computing self-attention relations brings more fine-grained self-attention knowledge and improves the performance of the student model.

% are computed via the multi-head dot-product, which is the same as self-attention distributions (dot-product of queries and keys).
% But the head number used in relations can be different from the attention head number of teacher and student models.
% For teacher and student models which uses different number of attention heads, we concatenate self-attention vectors (queries, keys and values) of different attention heads and then split them into vectors of the same number of relation heads to compute relations and perform KD training.
% Take queries as an example, we first concatenate queries of different attention heads and then split the concatenated vector into vectors of the number of relation heads.
% We also perform the same operation on values and keys, and 
We use $\mathbf{A}_1,\mathbf{A}_2,\mathbf{A}_3$ to denote the queries, keys and values of multiple relation heads.
% The number of relation heads can be different from the attention head number of teacher and student models, which allows arbitrary number of attention heads for student models. 
% Given a triple of queries $\mathbf{Q}$, keys $\mathbf{K}$, and values $\mathbf{V}$, 
% We first concatenate queries, keys, and values of different heads and split the three concatenate vectors into any relation head number of vectors.
% $\mathbf{A}_1,\mathbf{A}_2,\mathbf{A}_3$ are used to denote the three split vectors. 
% Self-attention relations between different combinations of query, key and value are computed via the multi-head scaled dot-product like self-attention distributions (dot-product of queries and keys),
% but the head number used in relations can be more flexible. 
% It can be different from the attention head number of teacher and student models, and allows arbitrary attention head number for student models. 
% Moreover, we find that using a larger number of relation heads achieves better performance.
The KL-divergence between multi-head self-attention relations of the teacher and student is used as the training objective: 
\begin{gather}
\mathcal{L} = \sum_{i=1}^{3}{\sum_{j=1}^{3}{\alpha_{ij}\mathcal{L}_{ij}}} \\
\mathcal{L}_{ij} = \frac{1}{A_{r}|x|}{\sum_{a=1}^{A_{r}}{\sum_{t=1}^{|x|}{D_{KL}(\mathbf{R}^{T}_{ij,l,a,t} \parallel \mathbf{R}^{S}_{ij,m,a,t})}}} \\
\mathbf{R}_{ij,l,a}^T = \softmax(\frac{\mathbf{A}_{i,l,a}^{T} \mathbf{A}_{j,l,a}^{T\intercal}}{ \sqrt{d_r}})  \\
\mathbf{R}_{ij,m,a}^S = \softmax(\frac{\mathbf{A}_{i,m,a}^{S} \mathbf{A}_{j,m,a}^{S\intercal}}{ \sqrt{d_r^{\prime}}}) 
\end{gather}
where $\mathbf{A}_{i,l,a}^{T} \in \mathbb{R}^{|x| \times d_r}$ and $\mathbf{A}_{i,m,a}^{S} \in \mathbb{R}^{|x| \times d_r^{\prime}}$ ($i \in [1,3]$) are the queries, keys and values of a relation head of $l$-th teacher layer and $m$-th student layer. 
% $d_h$ and $d_h^{\prime}$ are the hidden size of teacher and student. 
$d_r$ and $d_r^{\prime}$ are the relation head size of teacher and student models. 
$\mathbf{R}^{T}_{ij,l} \in \mathbb{R}^{A_r \times |x| \times |x|}$ is the self-attention relation of $\mathbf{A}_{i,l}^{T}$ and $\mathbf{A}_{j,l}^{T}$ of teacher model.
$\mathbf{R}^{T}_{ij,l,a} \in \mathbb{R}^{|x| \times |x|}$ is the self-attention relation of a teacher's relation head.
$\mathbf{R}^{S}_{ij,m} \in \mathbb{R}^{A_r \times |x| \times |x|}$ is the self-attention relation of student model. 
For example, $\mathbf{R}^{T}_{11,l}$ represents teacher's Q-Q attention relation in Figure~\ref{fig:method}.
$A_r$ is the number of relation heads.
If the number of relation heads and attention heads is the same, the Q-K relation is equivalent to the attention weights in self-attention module. 
% $d_r\times A_r$ and $d_r^{\prime}\times A_r$ are equal to the hidden size of teacher and student models.
% Experiments show that using a larger number of relation heads achieves better performance.
$\alpha_{ij} \in \{0, 1\}$ is the weight assigned to each self-attention relation loss. 
% We find that only using the query-query, key-key and value-value relations achieves competitive performance.
We transfer query-query, key-key and value-value self-attention relations to balance the performance and training cost.

\subsection{Layer Selection of Teacher Model}

Besides the knowledge used for distillation, mapping function between teacher and student layers is another key factor. 
As in \minilmvone{}, we only transfer the self-attention knowledge of one of the teacher layers to the student last layer. 
Only distilling one layer of the teacher model is fast and effective.
Different from previous work which usually conducts experiments on base-size teachers, we experiment with different large-size teachers and find that transferring self-attention knowledge of an upper middle layer performs better than using other layers.
For \bertlarge{} and \bertlargewwm{}, transferring the $21$-th (start at one) layer achieves the best performance. For \robertalarge{} and \xlmrlarge{}, using the self-attention knowledge of $19$-th layer achieves better performance.
For the base-size teacher, we also find that using teacher's last layer performs better.
% experiments indicate that using teacher's last layer performs better.
% than other layers.

% Most of previous work which conduct experiments on base-size model. We experiment with different large-size teachers and find that transferring a upper middle layer achieves better performance than other layers. Moreover, the best layer for different teachers is also different. For \bertlarge{} and \bertlargewwm{}, we find that transferring the 21-th layer achieves the best performance. For \robertalarge{}, using the 19-th layer is better.

\begin{table*}[t]
\begin{center}
\scalebox{0.675}{
\begin{tabular}{l|c|c|c|cccccccc|c}
\toprule
\bf Model & \bf Teacher & \bf \#Param & \bf SQuAD2 & \textbf{MNLI} & \textbf{QNLI} & \textbf{QQP} & \textbf{RTE}  & \textbf{SST} & \textbf{MRPC}  & \textbf{CoLA}  & \textbf{STS}  & \textbf{Avg} \\ \midrule
\bertbase{}     & - & 109M &  76.8           &     84.5      &      91.7      &     91.3      &     68.6      &     93.2      &     87.3      &     58.9      &     89.5     &     82.4      \\
\robertabase{}     & - & 125M &    83.7             &     87.6      &      92.8      &     \textbf{91.9}      &  78.7 &     94.8      & 90.2 & 63.6 &     91.2     &     86.1      \\
% \vtwobase{}      & - & 109M &   86.1     & 88.5 & 93.5  & 91.7 & 81.3      & 95.1 & \textbf{91.8} & \textbf{65.2} & 91.0     &     87.1 \\
\midrule
$12$$\times$$768$ \ours{}      & \bertlarge{} & 109M & 81.8 & 86.5 & 92.6 & 91.6 & 76.4 & 93.3 & 89.2 & 62.3 & 90.5 & 84.9 \\
% \ours{}      & \bertlargewwm{} & 109M & - & - & - & - & - & - & - & - & - & - \\ 
$12$$\times$$768$ \ours{}      & \robertalarge{} & 125M & \textbf{86.6} & \textbf{89.4} & \textbf{94.0} & 91.8 & \textbf{83.1} & \textbf{95.9} & \textbf{91.2} & \textbf{65.0} & \textbf{91.3}  & \textbf{87.6} \\
\bottomrule
\end{tabular}
}
\end{center}
\caption{
Results of our $12$$\times$$768$ models on the dev sets of GLUE benchmark and SQuAD 2.0. The fine-tuning results are an average of $4$ runs for each task. We report F1 for SQuAD 2.0, Pearson correlation for STS-B, Matthews correlation coefficient for CoLA and accuracy for the rest.
% We report F1 for SQuAD 2.0, Matthews correlation coefficient for CoLA, Pearson correlation coefficient for STS, and accuracy for the rest.
% Metrics of \ours{} are averaged over four runs for the tasks.
}
\label{tbl:glue:base}
\end{table*}

\begin{table}[t]
% \vskip 0.1in
\begin{center}
\scalebox{0.65}{
\begin{tabular}{l|ccc|c}
\toprule
\bf Model & \bf SQuAD2 & \bf MNLI-m & \bf SST-2 & \bf Avg \\ 
\midrule
% \multicolumn{5}{l}{\emph{12$\times$384 student model with 128 embedding size}} \\
\minilmvone{} (Last Layer) & 79.1 & 84.7 & 91.2 & 85.0 \\
\quad + Upper Middle Layer & 80.3 & 85.2 & 91.5 & 85.7 \\
% \quad + Layer-Wise Distillation & 78.2  & 83.5 & 90.8 & 84.2 \\
\midrule
$12$$\times$$384$ \ours{} & \textbf{80.7}  & \textbf{85.7} & \textbf{92.3} & \textbf{86.2} \\
\bottomrule
\end{tabular}
}
\end{center}
\caption{
Comparison of different methods using \bertlargewwm{} as the teacher. 
We report dev results of $12$$\times$$384$ student model with $128$ embedding size.
% Dev results of different methods using \bertlargewwm{} as the teacher. 
% Comparison between the publicly released $6$-layer models with 768 hidden size distilled from \bertbase{}.
% We compare task-agnostic distilled models without task-specific distillation and data augmentation.
% We report F1 for SQuAD 2.0, Matthews correlation coefficient for CoLA, and accuracy for other datasets. 
% The GLUE results of DistilBERT are taken from~\citet{distillbert}. 
% The rest results of DistilBERT, TinyBERT\footnotemark, \bertsmall{}, Truncated \bertbase{} and 6$\times$768 \minilmvone{} are taken from~\citet{minilmv1}.
% \bertsmall{}~\cite{studentInit} is trained using the MLM objective, without using knowledge distillation.
% We also report the results of truncated \bertbase{} and truncated \robertabase{}, which drops the top 6 layers of the base model.
% Top-layer dropping has been proven to be a strong baseline~\cite{poormanbert}. 
% The fine-tuning results are an average of $4$ runs.
% We report the SQuAD 2.0 result by fine-tuning their released model$^{5}$.
% For TinyBERT, we fine-tune the latest version of their public model$^{6}$ for a fair comparison. We also report the fine-tuning results of Truncated \bertbase{} and the $6$x$768$ BERT model (\bertsmall{})~\cite{studentInit} trained using the MLM objective as baselines. \citet{poormanbert} show that top-layer dropping consistently outperforms other strategies when
% dropping 6 layers, so we drop top 6 layers from \bertbase{} for truncated \bertbase{}. 
}
% \vskip -0.1in
\label{tbl:other_methods_exps}
\end{table}

\begin{table*}[t]
% \vskip 0.1in
\begin{center}
\scalebox{0.6}{
\begin{tabular}{l|c|ccc|ccccccccccccccc|c}
\toprule
\bf Model & \bf Teacher & \bf \#L & \bf \#H & \bf \#Param & \bf en & \bf fr & \bf es & \bf de & \bf el & \bf bg & \bf ru & \bf tr & \bf ar & \bf vi & \bf th & \bf zh & \bf hi & \bf sw & \bf ur & \bf Avg \\ 
\midrule
mBERT & - & 12 & 768 & 170M & 82.1 & 73.8 & 74.3 & 71.1 & 66.4 & 68.9 & 69.0 & 61.6 & 64.9 & 69.5 & 55.8 & 69.3 & 60.0 & 50.4 & 58.0 & 66.3 \\
XLM-$100$ & - & 16 & 1280 & 570M & 83.2 & 76.7 & 77.7 & 74.0 & 72.7 & 74.1 & 72.7 & 68.7 & 68.6 & 72.9 & 68.9 & 72.5 & 65.6 & 58.2 & 62.4 & 70.7 \\
% \xlmrbase{} & - & 12 & 768 & 84.6  & 78.4 & 78.9 & 76.8 & 75.9 & 77.3 & 75.4 & 73.2 & 71.5 & 75.4 & 72.5 & 74.9 & 71.1 & 65.2 & 66.5 & 74.5 \\
\xlmrbase{} & - & 12 & 768 & 277M & 85.8  & 79.7 & 80.7 & 78.7 & 77.5 & 79.6 & 78.1 & 74.2 & 73.8 & 76.5 & 74.6 & 76.7 & 72.4 & 66.5 & 68.3 & 76.2 \\
% \ours{}$^{a}$ & 12 & 384 & 81.5  & 74.8 & 75.7 & 72.9 & 73.0 & 74.5 & 71.3 & 69.7 & 68.8 & 72.1 & 67.8 & 70.0 & 66.2 & 63.3 & 64.2 & 71.1 \\
\midrule
\minilmvone{} & \xlmrbase{} & 6 & 384 & 107M & 79.2  & 72.3 & 73.1 & 70.3 & 69.1 & 72.0 & 69.1 & 64.5 & 64.9 & 69.0 & 66.0 & 67.8 & 62.9 & 59.0 & 60.6 & 68.0 \\
$6$$\times$$384$ \ours{} & \xlmrbase{} & 6 & 384 & 107M & 78.8  & 72.1 & 73.4 & 69.4 & 70.7 & 72.2 & 69.9 & 66.7 & 66.7 & 69.5 & 67.4 & 68.6 & 65.3 & 61.2 & 61.9 & 68.9 \\
$6$$\times$$384$ \ours{} & \xlmrlarge{} & 6 & 384 & 107M & 78.9  & 72.5 & 73.5 & 69.8 & 70.9 & 72.3 & 69.7 & 67.5 & 67.1 & 70.2 & 67.4 & 69.1 & 65.4 & 62.3 & 63.4 & \textbf{69.3} \\
\midrule
\minilmvone{} & \xlmrbase{} & 12 & 384 & 117M & 81.5  & 74.8 & 75.7 & 72.9 & 73.0 & 74.5 & 71.3 & 69.7 & 68.8 & 72.1 & 67.8 & 70.0 & 66.2 & 63.3 & 64.2 & 71.1 \\
$12$$\times$$384$ \ours{} & \xlmrbase{} & 12 & 384 & 117M & 82.2  & 75.3 & 76.5 & 73.3 & 73.8 & 75.1 & 73.1 & 70.7 & 69.5 & 72.8 & 69.4 & 71.5 & 68.6 & 65.6 & 65.5 & 72.2 \\
$12$$\times$$384$ \ours{} & \xlmrlarge{} & 12 & 384 & 117M & 82.4  & 76.2 & 76.8 & 73.9 & 74.6 & 75.5 & 73.7 & 70.8 & 70.5 & 73.4 & 71.1 & 72.5 & 68.8 & 66.8 & 67.0 & \textbf{72.9} \\
\bottomrule
\end{tabular}
}
\end{center}
\caption{
Cross-lingual classification results of our multilingual models on XNLI. We report the accuracy on each of the $15$ XNLI languages and the average accuracy. \#L and \#H indicate the number of layers and hidden size. 
% Results of mBERT are taken from~\citet{wu19mberteffective}. 
% Results of mBERT, XLM-100 and \xlmrbase{} are from~\citet{xlmr}. 
% Our fine-tuning results are averaged over $4$ runs.
}
% \vskip -0.1in
\label{tbl:xnli_exps}
\end{table*}

\begin{table*}[t]
% \vskip 0.1in
\begin{center}
\scalebox{0.61}{
\begin{tabular}{l|c|ccc|ccccccc|c}
\toprule
\bf Model & \bf Teacher & \bf \#L & \bf \#H  & \bf \#Param & \bf en & \bf es & \bf de & \bf ar & \bf hi & \bf vi & \bf zh & \bf Avg \\ 
\midrule
mBERT & - & 12 & 768 & 170M & 77.7 / 65.2 & 64.3 / 46.6 &  57.9 / 44.3 & 45.7 / 29.8 &  43.8 / 29.7 & 57.1 / 38.6 & 57.5 / 37.3 & 57.7 / 41.6 \\
XLM-$15$ & - & 12 & 1024 & 248M & 74.9 / 62.4 & 68.0 / 49.8 & 62.2 / 47.6 & 54.8 / 36.3 & 48.8 / 27.3 &  61.4 / 41.8 & 61.1 / 39.6 & 61.6 / 43.5 \\
% \xlmrbase{} & - & 12 & 768 & 77.8 / 65.3  & 67.2 / 49.7 & 60.8 / 47.1 & 53.0 / 34.7 & 57.9 / 41.7 & 63.1 / 43.1 & 60.2 / 38.0 & 62.9 / 45.7 \\
\xlmrbase{} & - & 12 & 768 & 277M & 77.1 / 64.6  & 67.4 / 49.6 & 60.9 / 46.7 & 54.9 / 36.6 & 59.4 / 42.9 & 64.5 / 44.7 & 61.8 / 39.3 & 63.7 / 46.3 \\
\xlmrbase{}$\dag$ & - & 12 & 768 & 277M & 80.3 / 67.4  & 67.0 / 49.2 & 62.7 / 48.3 & 55.0 / 35.6 & 60.4 / 43.7 & 66.5 / 45.9 & 62.3 / 38.3 & 64.9 / 46.9 \\
% \ours{}$^{a}$ & 12 & 384 & 79.4 / 66.5 & 66.1 / 47.5 &  61.2 / 46.5 & 54.9 / 34.9 &  58.5 / 41.3 & 63.1 / 42.1 & 59.0 / 33.8 & 63.2 / 44.7 \\
\midrule
\minilmvone{} & \xlmrbase{} & 6 & 384 & 107M & 75.5 / 61.9 & 55.6 / 38.2 &  53.3 / 37.7 & 43.5 / 26.2 &  46.9 / 31.5 & 52.0 / 33.1 & 48.8 / 27.3 & 53.7 / 36.6 \\
$6$$\times$$384$ \ours{} & \xlmrbase{} & 6 & 384 & 107M & 75.6 / 61.9 & 60.4 / 42.6 &  57.5 / 42.4 & 49.9 / 31.1 &  53.4 / 36.3 & 57.2 / 36.7 & 54.4 / 32.4 & 58.3 / 40.5 \\
$6$$\times$$384$ \ours{} & \xlmrlarge{} & 6 & 384 & 107M & 75.8 / 62.4 & 60.5 / 43.4 &  56.7 / 41.6 & 49.3 / 30.9 &  55.1 / 38.1 & 58.6 / 38.5 & 56.9 / 34.3 & \textbf{59.0} / \textbf{41.3} \\
\midrule
\minilmvone{} & \xlmrbase{} & 12 & 384 & 117M & 79.4 / 66.5 & 66.1 / 47.5 &  61.2 / 46.5 & 54.9 / 34.9 &  58.5 / 41.3 & 63.1 / 42.1 & 59.0 / 33.8 & 63.2 / 44.7 \\
$12$$\times$$384$ \ours{} & \xlmrbase{} & 12 & 384 & 117M & 79.0 / 65.8 & 66.6 / 48.9 &  62.1 / 47.4 & 54.4 / 35.5 &  60.6 / 43.6 & 64.7 / 44.1 & 59.4 / 36.5 & 63.8 / 46.0 \\
$12$$\times$$384$ \ours{} & \xlmrlarge{} & 12 & 384 & 117M & 78.9 / 65.9 & 67.3 / 49.4 &  61.9 / 47.0 & 56.0 / 36.5 &  63.0 / 45.9 & 65.1 / 44.4 & 61.9 / 39.2 & \textbf{64.9} / \textbf{46.9} \\
\bottomrule
\end{tabular}
}
\end{center}
% \vskip -0.1in
\caption{
Cross-lingual question answering results of our multilingual models on MLQA. We report the F1 and EM (exact match) scores on each of the $7$ MLQA languages. \#L and \#H indicate the number of layers and hidden size. $\dag$ indicates our fine-tuned results of \xlmrbase{}. 
% Results of mBERT and XLM-$15$ are taken from~\citet{mlqa}. $\dag$ indicates results of \xlmrbase{} taken from~\citet{xlmr}. 
% We also report our fine-tuned results ($\ddag$) of \xlmrbase{}. 
}
\label{tbl:mlqa_exps}
\end{table*}

\begin{table}[t]
% \vskip 0.1in
\begin{center}
\scalebox{0.65}{
\begin{tabular}{l|ccc|c}
\toprule
\bf Model & \bf SQuAD2 & \bf MNLI-m & \bf SST-2 & \bf Avg \\ 
\midrule
% \multicolumn{5}{l}{\emph{12$\times$384 student model with 128 embedding size}} \\
\ours{} (Q-Q + K-K + V-V) & \textbf{72.8} & \textbf{82.2} & \textbf{91.5} & \textbf{82.2} \\
\quad -- Q-Q Att-Rel & 71.6 & 81.9 & 90.6 & 81.4 \\
\quad -- K-K Att-Rel & 71.9 & 81.9 & 90.5 & 81.4 \\
\quad -- V-V Att-Rel & 71.5 & 81.6 & 90.5 & 81.2 \\
Q-K + V-V Att-Rels & 72.4 & \textbf{82.2} & 91.0 & 81.9 \\
\bottomrule
\end{tabular}
}
\end{center}
\caption{
Ablation studies of different self-attention relations. We report results of $6$$\times$$384$ student model distilled from \bertbase{}. The relation head number is $12$.
% Ablation studies. 
% Comparison between the publicly released $6$-layer models with 768 hidden size distilled from \bertbase{}.
% We compare task-agnostic distilled models without task-specific distillation and data augmentation.
% We report F1 for SQuAD 2.0, Matthews correlation coefficient for CoLA, and accuracy for other datasets. 
% The GLUE results of DistilBERT are taken from~\citet{distillbert}. 
% The rest results of DistilBERT, TinyBERT\footnotemark, \bertsmall{}, Truncated \bertbase{} and 6$\times$768 \minilmvone{} are taken from~\citet{minilmv1}.
% \bertsmall{}~\cite{studentInit} is trained using the MLM objective, without using knowledge distillation.
% We also report the results of truncated \bertbase{} and truncated \robertabase{}, which drops the top 6 layers of the base model.
% Top-layer dropping has been proven to be a strong baseline~\cite{poormanbert}. 
% The fine-tuning results are an average of $4$ runs.
% We report the SQuAD 2.0 result by fine-tuning their released model$^{5}$.
% For TinyBERT, we fine-tune the latest version of their public model$^{6}$ for a fair comparison. We also report the fine-tuning results of Truncated \bertbase{} and the $6$x$768$ BERT model (\bertsmall{})~\cite{studentInit} trained using the MLM objective as baselines. \citet{poormanbert} show that top-layer dropping consistently outperforms other strategies when
% dropping 6 layers, so we drop top 6 layers from \bertbase{} for truncated \bertbase{}. 
}
% \vskip -0.1in
\label{tbl:rels_abl_exps}
\end{table}

\begin{table}[t]
% \vskip 0.1in
\begin{center}
\scalebox{0.7}{
\begin{tabular}{lcccc}
\toprule
\#Relation Heads & 6 & 12 & 24 & \bf 48 \\ 
\midrule
\multicolumn{5}{l}{\emph{$6$$\times$$384$ model distilled from \robertabase{}}} \\
\quad MNLI-m & 82.8 & 82.9 & 83.0 & \textbf{83.4} \\ 
\quad SQuAD 2.0 & 74.5 & 75.0 & 74.9 & \textbf{75.7} \\ 
\midrule
\multicolumn{5}{l}{\emph{$6$$\times$$384$ model distilled from \bertbase{}}} \\
\quad MNLI-m & 81.9 & 82.2 & 82.2 & \textbf{82.4} \\ 
\quad SQuAD 2.0 & 71.9 & 72.8 & 72.7 & \textbf{73.0} \\ 
\bottomrule
\end{tabular}
}
\end{center}
\caption{
Results of $6$$\times$$384$ model using different number of relation heads. 
}
% \vskip -0.1in
\label{tbl:exps_rel_heads}
\end{table}

% \begin{table}[t]
% % \vskip 0.1in
% \centering
% % \small
% \scalebox{0.75}{
% \begin{tabular}{lclccclccc}
% \toprule
% \multirow{2}{*}{\bf Model}                          & \multirow{2}{*}{\bf \#Param} & & \multicolumn{3}{c}{\bf XSum} \\
% &                         & & RG-1  & RG-2  &     RG-L     \\ \midrule
% \textsc{Lead}-3 &                                 & & 16.30 & 1.60  &    11.95     \\
% \textsc{PtrNet} &             & & 28.10 & 8.02  &    21.72     \\
% \bartlarge{}                                        & 400M        & & 45.14 & 22.27 &    37.25     \\
% % \multicolumn{8}{l}{\textit{Fine-tuning \textsc{base}-size pre-trained models}} & & \\
% {MASS$_{\textsc{base}}$}                            & 123M       & & 39.75 & 17.24 &    31.95     \\
% \textsc{BERTSumAbs}                                 & 156M         & & 38.76 & 16.33 &    31.15     \\
% \midrule
% 6$\times$384 \minilmvone{}                             & 22M     & & 38.79 & 16.39 &    31.10     \\
% 6$\times$384 \ours{}                                   & 30M     & & 39.00 & 16.45 &    31.46     \\
% \midrule
% 12$\times$384 \minilmvone{}                             & 33M   & & 40.43 & 17.72 &    32.60     \\
% 12$\times$384 \ours{}                                   & 41M   & & 41.22 & 18.22 &    33.44     \\
% \bottomrule
% \end{tabular}
% }
% % \vskip -0.1in
% \caption{
% Abstractive summarization results of our $12\times384$ and $6\times384$ models distilled from \robertalarge{} on XSum.
% The evaluation metric is the F1 version of ROUGE (RG) scores.
% }
% \label{tbl:summ_exps}
% \end{table}

\section{Experiments}

We conduct distillation experiments on different teacher models including \bertbase{}, \bertlarge{}, \bertlargewwm{}, \robertabase{}, \robertalarge{}, \xlmrbase{} and \xlmrlarge{}. 
% We use multi-head query-query, key-key and value-value relations to perform KD training.

\subsection{Setup}

We use the uncased version for three BERT teacher models. 
% \bertbase{} consists of 12 Transformer layers with 768 hidden size, and 12 attention heads. 
% It contains about 109M parameters. 
% \bertlarge{} and \bertlargewwm{} are a 24-layer Transformer with 1024 hidden size and 16 attention heads. 
% They are both consist of 340M parameters. 
For the pre-training data, we use English Wikipedia and BookCorpus~\cite{bookcorpus}.
% and follow the preprocess and the WordPiece tokenization of~\citet{bert}.  
We train student models using $256$ as the batch size and 6e-4 as the peak learning rate for $400,000$ steps. 
We use linear warmup over the first $4,000$ steps and linear decay.
We use Adam~\cite{adam} with $\beta_1=0.9$, $\beta_2=0.999$. The maximum sequence length is set to $512$. The dropout rate and weight decay are $0.1$ and $0.01$. 
The number of attention heads is $12$ for all student models.
% For \bertlarge{} and \bertlargewwm{}, we use the self-attention knowledge of the $21$-th layer to train the student model.
The number of relation heads is $48$ and $64$ for base-size and large-size teacher model, respectively.
The student models are initialized randomly.

% \footnote{\url{skylion007.github.io/OpenWebTextCorpus}}
For models distilled from RoBERTa, we use similar pre-training datasets as in~\citet{roberta}.
% including text corpora from English Wikipedia, BookCorpus~\cite{bookcorpus}, OpenWebText, CC-News~\cite{roberta}, and Stories~\cite{stories_data}. 
% We use the self-attention knowledge of teacher's $19$-th layer for training the small models.
For the $12$$\times$$768$ student model, we use Adam with $\beta_1=0.9$, $\beta_2=0.98$. The rest hyper-parameters are the same as models distilled from BERT.

For multilingual student models distilled from XLM-R, we perform training using the same datasets as in~\citet{xlmr} for $1,000,000$ steps. 
We conduct distillation experiments using $8$ V100 GPUs with mixed precision training.

\subsection{Downstream Tasks}

Following previous pre-training~\cite{bert,roberta,xlmr} and task-agnostic distillation~\cite{mobilebert,tinybert} work, we evaluate the English student models on GLUE benchmark and extractive question answering. The multilingual models are evaluated on cross-lingual natural language inference and cross-lingual question answering.

\paragraph{GLUE} 

General Language Understanding Evaluation (GLUE) benchmark~\cite{wang2018glue} consists of two single-sentence classification tasks (SST-2~\cite{sst2013} and CoLA~\cite{cola2018}), three similarity and paraphrase tasks (MRPC~\cite{mrpc2005}, STS-B~\cite{sts-b2017} and QQP), and four inference tasks (MNLI~\cite{mnli2017}, QNLI~\cite{squad1}, RTE~\cite{rte1,rte2,rte3,rte5} and WNLI~\cite{winograd2012}). 
% Following BERT~\cite{bert}, a task-specific linear layer is added on top of the \sptk{CLS} representation. 

\begin{table*}[t]
% \vskip 0.1in
\begin{center}
\scalebox{0.62}{
\begin{tabular}{l|c|cc|c|cccccccc|c}
\toprule
\bf Model & \bf Teacher & \bf \#Param & \bf Speedup & \bf SQuAD2 & \bf MNLI-m/mm & \bf QNLI & \bf QQP & \bf RTE & \bf SST & \bf MRPC & \bf CoLA & \bf STS & \bf Avg \\ 
\midrule
\bertbase{} & - & 109M & 1.0$\times$ & 76.8 & 84.6/83.4 & 90.5 & 71.2 & 66.4 & 93.5 & 88.9 & 52.1 & 85.8 & 79.3 \\
\midrule
MobileBERT & IB-\bertlarge{} & 25M & 1.8$\times$ & 80.2 & 84.3/83.4 & 91.6 & 70.5 & 70.4 & 92.6 & 88.8 & \textbf{51.1} & 84.8 & 79.8 \\
$12$$\times$$384$ \ours{} & \bertlargewwm{} & 25M & 2.7$\times$ & 80.7 & \textbf{85.9}/84.6 & 91.9 & 71.4 & 71.9 & 93.3 & 89.2 & 44.9 & 85.5 & 79.9 \\
\quad + More Att-Rels & \bertlargewwm{} & 25M & 2.7$\times$ & \textbf{80.9} & 85.8/\textbf{84.8} & \textbf{92.3} & \textbf{71.6} & \textbf{72.0} & \textbf{93.6} & \textbf{89.7} & 46.6 & \textbf{86.0} & \textbf{80.3} \\
\bottomrule
\end{tabular}
}
\end{center}
\caption{
Comparison between MobileBERT and the same-size model ($12$ layers, $384$ hidden size and $128$ embedding size) distilled form \bertlarge{} (Whole Word Masking) on GLUE test sets and SQuAD 2.0 dev set. Following MobileBERT~\cite{mobilebert}, the reported results are directly fine-tuned on downstream tasks. We compute the speedup of MobileBERT according to their reported latency.
}
% \vskip -0.1in
\label{tbl:12l_384_exps}
\end{table*}

\begin{table}[t]
% \vskip 0.1in
\begin{center}
\scalebox{0.62}{
\begin{tabular}{l|c|ccc}
\toprule
\bf Model & \bf Teacher & \bf SQuAD2 & \bf MNLI-m & \bf SST-2 \\ 
\midrule
% \multicolumn{5}{l}{\emph{$6$$\times$$384$ models distilled from different teachers}} \\
$6$$\times$$384$ Ours & \bertbase{} & 72.9 & \textbf{82.8} & 91.3 \\
\quad + More Att-Rels & \bertbase{} & \textbf{73.3} & \textbf{82.8} & \textbf{91.6} \\
\midrule
$6$$\times$$384$ Ours & \bertlarge{} & 74.3  & 83.0 & 91.1 \\
\quad + More Att-Rels & \bertlarge{} & \textbf{74.7}  & \textbf{83.2} & \textbf{92.4} \\
\midrule
$6$$\times$$384$ Ours & \robertalarge{} & \textbf{76.4}  & \textbf{84.4} & 92.0 \\
\quad + More Att-Rels & \robertalarge{} & 76.0  & \textbf{84.4} & \textbf{92.1} \\
\midrule
% \multicolumn{5}{l}{\emph{$6$$\times$$768$ models distilled from different teachers}} \\
$6$$\times$$768$ Ours & \bertbase{} & 76.3  & 84.2 & \textbf{92.4} \\
\quad + More Att-Rels & \bertbase{} & \textbf{76.8} & \textbf{84.4} & 92.3 \\
\midrule
$6$$\times$$768$ Ours & \bertlarge{} & 77.7  & 85.0 & \textbf{92.5} \\
\quad + More Att-Rels & \bertlarge{} & \textbf{78.1}  & \textbf{85.2} & \textbf{92.5} \\
\midrule
$6$$\times$$768$ Ours & \robertalarge{} & \textbf{81.6}  & 87.0 & \textbf{94.5} \\
\quad + More Att-Rels & \robertalarge{} & 81.2  & \textbf{87.3} & 94.1 \\
% \midrule
% 12$\times$384 Ours & \robertalarge{} & 33M & $\times$2.7 & \textbf{82.3}  & \textbf{86.9} & \textbf{92.7} & \textbf{91.3} & \textbf{77.6} & \textbf{93.5} & \textbf{90.0} & \textbf{53.8} & \textbf{83.5} \\
\bottomrule
\end{tabular}
}
\end{center}
\caption{
Results of introducing more self-attention relations (Q-K, K-Q, Q-V, V-Q, K-V and V-K relations). 
% Comparison between the publicly released $6$-layer models with 768 hidden size distilled from \bertbase{}.
% We compare task-agnostic distilled models without task-specific distillation and data augmentation.
% We report F1 for SQuAD 2.0, Matthews correlation coefficient for CoLA, and accuracy for other datasets. 
% The GLUE results of DistilBERT are taken from~\citet{distillbert}. 
% The rest results of DistilBERT, TinyBERT\footnotemark, \bertsmall{}, Truncated \bertbase{} and 6$\times$768 \minilmvone{} are taken from~\citet{minilmv1}.
% \bertsmall{}~\cite{studentInit} is trained using the MLM objective, without using knowledge distillation.
% We also report the results of truncated \bertbase{} and truncated \robertabase{}, which drops the top 6 layers of the base model.
% Top-layer dropping has been proven to be a strong baseline~\cite{poormanbert}. 
% The fine-tuning results are an average of $4$ runs.
% We report the SQuAD 2.0 result by fine-tuning their released model$^{5}$.
% For TinyBERT, we fine-tune the latest version of their public model$^{6}$ for a fair comparison. We also report the fine-tuning results of Truncated \bertbase{} and the $6$x$768$ BERT model (\bertsmall{})~\cite{studentInit} trained using the MLM objective as baselines. \citet{poormanbert} show that top-layer dropping consistently outperforms other strategies when
% dropping 6 layers, so we drop top 6 layers from \bertbase{} for truncated \bertbase{}. 
}
% \vskip -0.1in
\label{tbl:more_rels_exps}
\end{table}

\paragraph{Extractive Question Answering}

The task aims to predict a continuous sub-span of the passage to answer the question. 
% SQuAD 2.0~\cite{squad2} has been served as a major question answering benchmark.
We evaluate on SQuAD 2.0~\cite{squad2}, which has been served as a major question answering benchmark.
% We pack the question and passage tokens together with special tokens to form the input: ``\sptk{CLS} $Q$ \sptk{SEP} $P$ \sptk{SEP}". Two linear output layers are introduced to predict the probability of each token being the start and end positions of the answer span. The questions that do not have an answer are treated as having an answer span with start and end at the \sptk{CLS} token.

\paragraph{Cross-lingual Natural Language Inference (XNLI)} XNLI~\cite{xnli} is a cross-lingual classification benchmark. It aims to identity the semantic relationship between two sentences and provides instances in $15$ languages. 

\paragraph{Cross-lingual Question Answering} We use MLQA~\cite{mlqa} to evaluate multilingual models. MLQA extends English SQuAD dataset~\cite{squad1} to seven languages. 

\begin{figure*}
\centering
\begin{tabular}{@{}ccc@{}}
\includegraphics[width=4.8cm]{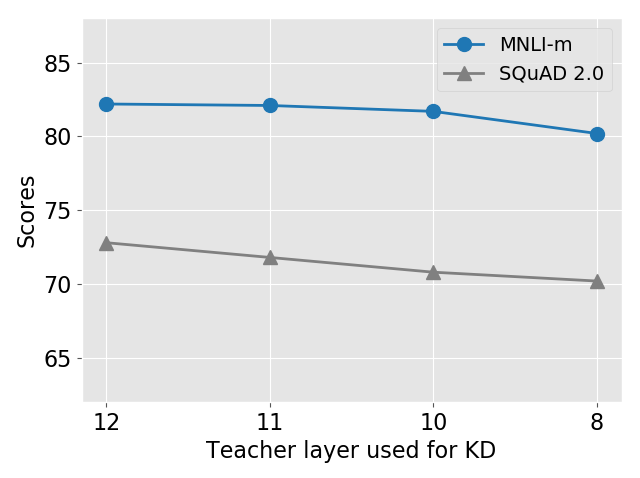} &
\includegraphics[width=4.8cm]{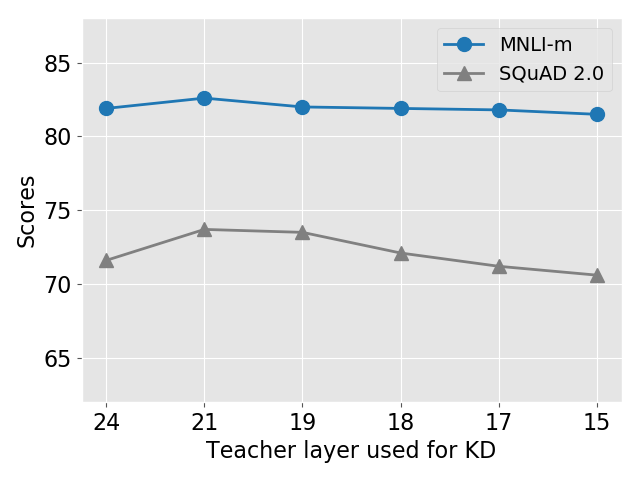} &
\includegraphics[width=4.8cm]{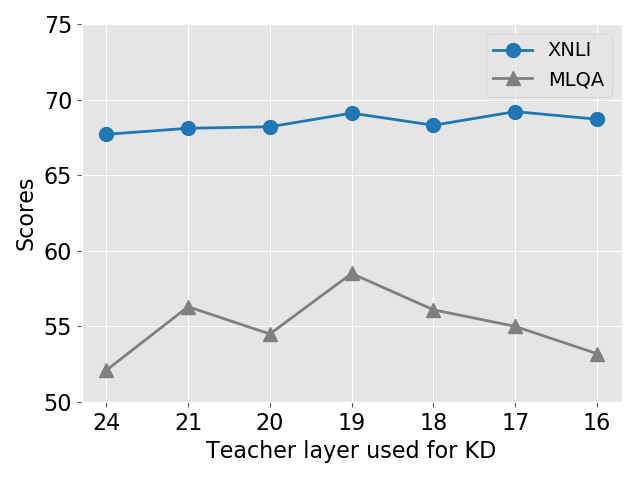} \\
\small
\shortstack{(a) \bertbase{} as the teacher}
&
\small
\shortstack{(b) \bertlarge{} as the teacher}
&
\small
\shortstack{(c) \xlmrlarge{} as the teacher} \\
\end{tabular}
\caption{
$6$$\times$$384$ models trained using different \bertbase{} (a), \bertlarge{} (b) and \xlmrlarge{} (c) layers.
}
\label{fig:analyze}
\end{figure*}

\subsection{Main Results}

Table~\ref{tbl:main_exps} presents the dev results of $6$$\times$$384$ and $6$$\times$$768$ models distilled from \bertbase{}, \bertlarge{} and \robertalarge{} on GLUE and SQuAD 2.0. 
(1) Previous methods~\cite{distillbert,tinybert,patientdistill,minilmv1} usually distill \bertbase{} into a $6$-layer model with $768$ hidden size. We first report results of the same setting.
% our 6$\times$768 model using the same teacher. 
Our $6$$\times$$768$ model outperforms DistilBERT, TinyBERT, \minilmvone{} and two BERT baselines across most tasks. Moreover, our method allows more flexibility for the number of attention heads of student models. (2) Both $6$$\times$$384$ and $6$$\times$$768$ models distilled from \bertlarge{} outperform models distilled from \bertbase{}. The $6$$\times$$768$ model distilled from \bertlarge{} is $2.0\times$ faster than \bertbase{}, while achieving better performance. 
(3) Student models distilled from \robertalarge{} achieve further improvements. Better teacher results in better students. 
Multi-head self-attention relation distillation is effective for different large-size pretrained Transformers.

We report the results of $6$$\times$$768$ students distilled from \bertbase{} and \bertlarge{} on GLUE test sets and SQuAD 2.0 dev set in Table~\ref{tbl:6l_768_test_exps}. $6$$\times$$768$ model distilled from \bertbase{} retains more than $99\%$ accuracy of its teacher while using $50\%$ Transformer parameters. $6$$\times$$768$ model distilled from \bertlarge{} compares favorably with \bertbase{}.

We compress \robertalarge{} and \bertlarge{} into a base-size student model. 
% The base model is trained using 256 as the batch size for $400,000$ steps. 
Dev results of GLUE benchmark and SQuAD 2.0 are shown in Table~\ref{tbl:glue:base}.
Our base-size models distilled from large-size teacher outperforms \bertbase{} and \robertabase{}.
Our method can be adopted to train students in different parameter size.
% Our method can also be employed to train a base-size model.
Moreover, our student distilled from \robertalarge{} uses a much smaller (almost $32\times$ smaller) training batch size and fewer training steps than \robertabase{}. 
% The computation cost is much smaller.
Our method requires much fewer training examples.
% and has a lower computation cost.

Most of previous work conducts experiments using base-size teachers.
To compare with previous methods on large-size teacher, we reimplement \minilmvone{} and compress \bertlargewwm{} into a $12$$\times$$384$ student model.
Dev results of SQuAD 2.0, MNLI-m and SST-2 are presented in Table~\ref{tbl:other_methods_exps}.
Our method also outperforms \minilmvone{} for large-size teachers.
Moreover, we report results of distilling an upper middle layer instead of the last layer for \minilmvone{}. 
Layer selection is also effective for \minilmvone{} when distilling large-size teachers.

Table~\ref{tbl:xnli_exps} and Table~\ref{tbl:mlqa_exps} show the results of our student models distilled from XLM-R on XNLI and MLQA. 
For XNLI, the best single model is selected on the joint dev set of all the languages as in~\citet{xlmr}.
Following~\citet{mlqa}, we adopt SQuAD
1.1 as training data and evaluate on MLQA English development set for early stopping. 
Our $6$$\times$$384$ model outperforms mBERT~\cite{bert} with $5.3\times$ speedup. 
Our method also performs better than \minilmvone{}, which further validates the effectiveness of multi-head self-attention relation distillation. 
Transferring multi-head self-attention relations can bring more fine-grained self-attention knowledge.

% \begin{figure}[t]
% \centering
% \includegraphics[width=6cm]{layer_analysis_bertbase_linechart.png}
% \caption{Results of 6$\times$384 model (12 attention heads, 12 relation heads) trained using different \bertbase{} layers.}
% \label{fig:analysis_bertbase_layer}
% \end{figure}

% \begin{figure}[t]
% \centering
% \includegraphics[width=6cm]{layer_analysis_bertlarge_linechart.png}
% \caption{Results of 6$\times$384 model (12 attention heads, 64 relation heads) trained using different \xlmrlarge{} layers.}
% \label{fig:analysis_bertlarge_layer}
% \end{figure}

% \begin{figure}[t]
% \centering
% \includegraphics[width=6cm]{layer_analysis_xlmrlarge_linechart.png}
% \caption{Results of 6$\times$384 model (6 attention heads, 16 relation heads) trained using different \bertlarge{} layers.}
% \label{fig:analysis_xlmrlarge_layer}
% \end{figure}

\subsection{Ablation Studies}

\paragraph{Effect of using different self-attention relations}

We perform ablation studies to analyse the contribution of different self-attention relations. 
Dev results of three tasks are illustrated in Table~\ref{tbl:rels_abl_exps}. 
Q-Q, K-K and V-V self-attention relations positively contribute to the final results.
Besides, we also compare Q-Q + K-K + V-V with Q-K + V-V given queries and keys are employed to compute self-attention distributions in self-attention module.
Experimental result shows that using Q-Q + K-K + V-V achieves better performance.

% Figure~\ref{fig:analysis_bertbase_layer} and ~\ref{fig:analysis_bertlarge_layer} present the results of 6$\times$384 model distilled from different layers of \bertbase{} and \bertlarge{}.
% For \bertbase{}, using the last layer achieves better performance than other layers. 
% For \bertlarge{}, we find that using one of the upper middle layers achieves the best performance. 
% The same trend is also observed for \bertlargewwm{} and \robertalarge{}. 

\paragraph{Effect of distilling different teacher layers} Figure~\ref{fig:analyze} presents the results of $6$$\times$$384$ model distilled from different layers of \bertbase{}, \bertlarge{} and \xlmrlarge{}.
For \bertbase{}, using the last layer achieves better performance than other layers. 
For \bertlarge{} and \xlmrlarge{}, we find that using one of the upper middle layers achieves the best performance. 
The same trend is also observed for \bertlargewwm{} and \robertalarge{}. 

\paragraph{Effect of different number of relation heads} Table~\ref{tbl:exps_rel_heads} shows the results of $6$$\times$$384$ model distilled from \bertbase{} and \robertabase{} using different number of relation heads. 
Using a larger number of relation heads achieves better performance.
More fine-grained self-attention knowledge can be captured by using more relation heads, which helps the student to deeply mimic the self-attention module of its teacher. 
Besides, we find that the number of relation heads is not required to be a positive multiple of both the number of student and teacher attention heads. The relation head can be a fragment of a single attention head or contains fragments from multiple attention heads.

\section{Discussion}

\subsection{Comparison with MobileBERT}
MobileBERT~\cite{mobilebert} compresses a specially designed teacher model (in the \bertlarge{} size) with inverted bottleneck modules into a $24$-layer student using the bottleneck modules. 
Since our goal is to compress different large models (e.g. BERT and RoBERTa) to small models using standard Transformer architecture, we note that our student model can not directly compare with MobileBERT.
We provide results of a student model with the same parameter size for a reference.
A public large-size model (\bertlargewwm{}) is used as the teacher, which achieves similar performance as MobileBERT's teacher. 
We distill \bertlargewwm{} into a student model ($25$M parameters) using the same training data (i.e., English Wikipedia and BookCorpus).
The test results of GLUE and dev result of SQuAD 2.0 are illustrated in Table~\ref{tbl:12l_384_exps}. 
Our model outperforms MobileBERT across most tasks with a faster inference speed. 
Moreover, our method can be applied for different teachers and has much fewer restriction of students.

We also observe that our model performs relatively worse on CoLA compared with MobileBERT. The task of CoLA is to evaluate the grammatical acceptability of a sentence. It requires more fine-grained linguistic knowledge that can be learnt from language modeling objectives. 
% This has been verified in MobileBERT that fine-tuning the model using the MLM objective brings improvements for CoLA. However, our preliminary experiment results show that this strategy will improve the results on CoLA while drop on other GLUE tasks.
Fine-tuning the model using the MLM objective as in MobileBERT brings improvements for CoLA. However, our preliminary experiments show that this strategy will lead to slight drop for other GLUE tasks.

% Since this work focuses on a general distillation approach for different teachers and different sized students, we note that our student model can not directly compare with MobileBERT. 

% To compare with MobileBERT, we use a public large-size model (\bertlargewwm{}) as the teacher, which achieves similar performance as the teacher of MobileBERT. We distill \bertlargewwm{} into a student model, which contains the same number of parameters (25M parameters, 12$\times$384 with 128 embedding size), using the same training data (i.e., English Wikipedia and BookCorpus). The test results of GLUE benchmark and dev result of SQuAD 2.0 are illustrated in Table~\ref{tbl:12l_384_exps}. \ours{} outperforms MobileBERT across most tasks with a faster inference speed. 
% Moreover, our method can be applied for different teachers and has much fewer restrictions of student models.

\subsection{Results of More Self-Attention Relations}

In Table~\ref{tbl:12l_384_exps} and~\ref{tbl:more_rels_exps}, we report results of students trained using more self-attention relations (Q-K, K-Q, Q-V, V-Q, K-V and V-K relations). We observe improvements across most tasks, especially for student models distilled from BERT. 
Fine-grained self-attention knowledge in more attention relations improves our students.
However, introducing more self-attention relations also brings a higher computational cost. In order to achieve a balance between performance and computational cost, we choose to transfer Q-Q, K-K and V-V self-attention relations instead of all self-attention relations in this work. 

\section{Conclusion}

% In this work, we present a simple and effective approach for compressing pretrained Transformers.
We generalize deep self-attention distillation in \minilmvone{} by employing multi-head self-attention relations to train the student.
% to deeply mimic the self-attention module of its teacher.
Our method introduces more fine-grained self-attention knowledge and eliminates the restriction of the number of student's attention heads.
% , which is required to be the same as its teacher for previous work transferring self-attention distributions. 
% introduced by transferring teacher's self-attention distributions as in existing work.
% Using multi-head self-attention relations eliminates the restriction on the number of attention heads introduced by transferring teacher's self-attention distributions as in existing work and capture more fine-grained self-attention knowledge.
Moreover, we show that transferring the self-attention knowledge of an upper middle layer achieves better performance for large-size teachers.
Our monolingual and multilingual models distilled from BERT, RoBERTa and XLM-R obtain competitive performance and outperform state-of-the-art methods.
For future work, we are exploring an automatic layer selection algorithm. We also would like to apply our method to larger pretrained Transformers.

\bibliography{anthology,minilmv2}
\bibliographystyle{acl_natbib}

\appendix

\section{GLUE Benchmark}

The summary of datasets used for the General Language Understanding Evaluation (GLUE) benchmark\footnote{\url{https://gluebenchmark.com/}}~\cite{wang2018glue} is presented in Table~\ref{tbl:glue:datasets}.

\begin{table}[t]
% \vskip 0.1in
\centering
\scalebox{0.72}{
\begin{tabular}{lcccc}
\toprule \textbf{Corpus} & \textbf{\#Train} & \textbf{\#Dev} & \textbf{\#Test} & \textbf{Metrics}   \\ \midrule
\multicolumn{5}{l}{\emph{Single-Sentence Tasks}} \\
CoLA & 8.5k & 1k & 1k & Matthews Corr \\
SST-2 & 67k & 872 & 1.8k & Accuracy \\ 
\midrule
\multicolumn{5}{l}{\emph{Similarity and Paraphrase Tasks}} \\
QQP & 364k & 40k & 391k & Accuracy/F1 \\ 
MRPC & 3.7k & 408 & 1.7k & Accuracy/F1\\ 
STS-B & 7k & 1.5k & 1.4k & Pearson/Spearman Corr \\ 
\midrule
\multicolumn{5}{l}{\emph{Inference Tasks}} \\
MNLI & 393k & 20k &20k & Accuracy\\
RTE &2.5k & 276 & 3k & Accuracy \\ 
QNLI & 105k & 5.5k & 5.5k & Accuracy\\
WNLI & 634 & 71 & 146 & Accuracy\\ 
\bottomrule
\end{tabular}
}
\caption{Summary of the GLUE benchmark.}
% \vskip -0.1in
\label{tbl:glue:datasets}
\end{table}

\begin{table}[t]
% \vskip 0.1in
\centering
\small
\begin{tabular}{cccc}
\toprule
\textbf{\#Train} & \textbf{\#Dev} & \textbf{\#Test} & \textbf{Metrics}   \\ \midrule
130,319 & 11,873 & 8,862 & Exact Match/F1 \\
\bottomrule
\end{tabular}
% \vskip -0.1in
\caption{Dataset statistics and metrics of SQuAD 2.0.}
\label{tbl:squad2}
\end{table}

\section{SQuAD 2.0}

We present the dataset statistics and metrics of SQuAD 2.0\footnote{\url{http://stanford-qa.com}}~\cite{squad2} in Table~\ref{tbl:squad2}.

\section{Hyper-parameters for Fine-tuning}

\paragraph{Extractive Question Answering} For SQuAD 2.0, the maximum sequence length is $384$. 
The batch size is set to $32$. 
We choose learning rates from \{3e-5, 6e-5, 8e-5, 9e-5\} and fine-tune the model for 3 epochs.
The warmup ration and weight decay is 0.1 and 0.01.

\paragraph{GLUE} The maximum sequence length is $128$ for the GLUE benchmark. We set batch size to $32$, choose learning rates from \{1e-5, 1.5e-5, 2e-5, 3e-5, 5e-5\} and epochs from \{$3$, $5$, $10$\} for different student models. We fine-tune CoLA task with longer training steps (25 epochs). The warmup ration and weight decay is 0.1 and 0.01.

\paragraph{Cross-lingual Natural Language Inference (XNLI)} The maximum sequence length is $256$ for XNLI. We fine-tune $10$ epochs using $64$ as the batch size. The learning rates are chosen from \{3e-5, 4e-5, 5e-5, 6e-5\}.

\paragraph{Cross-lingual Question Answering} For MLQA, the maximum sequence length is $512$. We fine-tune $4$ epochs using $32$ as the batch size. The learning rates are chosen from \{3e-5, 4e-5, 5e-5, 6e-5\}.

\end{document}